# Validation of image-guided cochlear implant programming techniques

Yiyuan Zhao[a], Jianing Wang[a], Rui Li[a], Robert F. Labadie[b], Benoit M. Dawant[a], and Jack H. Noble[a]

[a] Department of Electrical Engineering and Computer Science, Vanderbilt University, Nashville, TN 37235, USA

[b] Department of Otolaryngology – Head & Neck Surgery, Vanderbilt University Medical Center, Nashville, TN 37235, USA

**Corresponding Author**

Jack H. Noble

Electrical Engineering and Computer Science, Vanderbilt University,

2301 Vanderbilt Pl., VU Station B #351679

Nashville, TN 37235

**Work**: (615)-875-5539

**Email**: jack.noble@vanderbilt.edu

**Abstract**

Cochlear implants (CIs) are a standard treatment for patients who experience severe to profound hearing loss. Recent studies have shown that hearing outcome is correlated with intra-cochlear anatomy and electrode placement. Our group has developed image-guided CI programming (IGCIP) techniques that use image analysis methods to both segment the inner ear structures in pre- or post-implantation CT images and localize the CI electrodes in post-implantation CT images. This permits to assist audiologists with CI programming by suggesting which among the contacts should be deactivated to reduce electrode interaction that is known to affect outcomes. Clinical studies have shown that IGCIP can improve hearing outcomes for CI recipients. However, the sensitivity of IGCIP with respect to the accuracy of the two major steps: electrode localization and intra-cochlear anatomy segmentation, is unknown. In this article, we create a ground truth dataset with conventional CT and µCT images of 35 temporal bone specimens to both rigorously characterize the accuracy of these two steps and assess how inaccuracies in these steps affect the overall results. Our study results show that when clinical pre- and post-implantation CTs are available, IGCIP produces results that are comparable to those obtained with the corresponding ground truth in 86.7% of the subjects tested. When only post-implantation CTs are available, this number is 83.3%. These results suggest that our current method is robust to errors in segmentation and localization but also that it can be improved upon.

**Keywords:** cochlear implant, ground truth, segmentation, validation

1. **Introduction**

Cochlear implants (CIs) are neural prosthetic devices that are the standard of care treatment for patients experiencing severe to profound hearing loss [1]. The external components of a CI device include a microphone, a signal processor, and a signal transmitter, which are used to receive and process sounds, and send signals to implanted CI electrodes. The major internal component is the implanted CI electrode array. The implanted CI electrodes bypass the damaged cochlea and directly stimulate the auditory nerves to induce a sense of hearing for the recipient. During CI surgery, a surgeon threads a CI electrode array into a recipient's cochlea. After the surgery, an audiologist needs to program the CI device which includes determining a series of CI instructions. The programming procedure involves specifying the stimulation levels for each electrode based on the recipient's perceived loudness, and the selection of a frequency allocation table, which determines which electrode is to be activated when a specific frequency is detected in the incoming sound [2]. CIs lead to remarkable success in hearing restoration among the majority of recipients [3-4]. However, there are still a significant number of CI recipients experiencing only marginal benefit.

Recent studies have indicated that hearing outcomes with CI devices are correlated with the intra-cochlear locations of CI electrodes [5-10]. As the electrode array is blindly inserted by a surgeon, the intra-cochlear locations of CI electrodes are generally unknown. Thus, audiologists do not have information about locations of CI electrodes with respect to the auditory nerves. In the traditional CI programming procedure, the audiologist assumes the electrodes are optimally situated and selects a default frequency allocation table. This may lead to an artifact named "electrode interaction" [11-12], as shown in Figure 1 as overlapping stimulation of electrodes. Electrode interaction occurs when multiple CI electrodes are stimulating the same group of auditory nerves. In natural hearing, a specific group of nerves are activated in response to a specific frequency band. In a CI-assisted hearing process with electrode interaction, the same nerve group is activated in response to multiple frequency bands, which is thought to create spectral smearing and negatively affect hearing outcomes. It is possible to alleviate the negative effect of electrode interaction by selecting a subset of the available electrodes to keep active, aka the "electrode configuration", that do not have overlapping stimulation patterns. However, without the benefit of knowing the spatial relationship between the electrodes and the auditory neural sites, selecting such an electrode configuration is not possible and audiologists typically leave active all available electrodes.

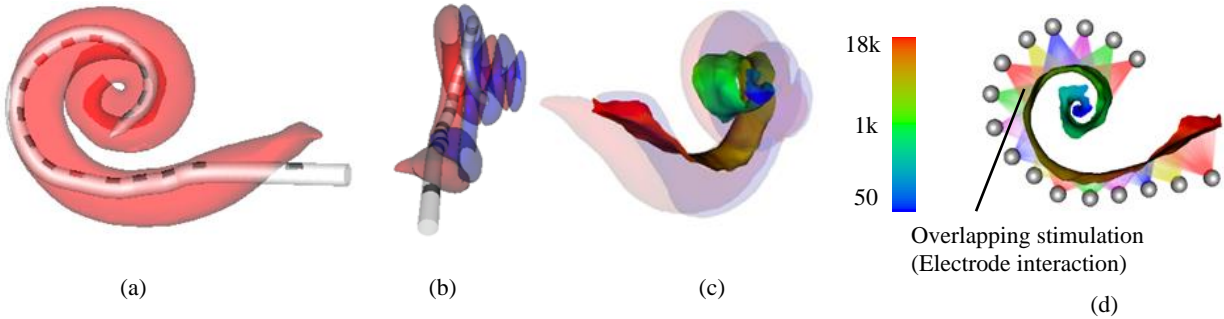

**Figure 1.** Panels (a) and (b) show a CI electrode array superimposed on the scala tympani (red) and scala vestibuli (blue) cavities of the cochlea in posterior-to-anterior and lateral-to-medial views, respectively. Panel (c) shows the scalae and neural activation region color-coded by place frequency in Hz. Panel (d) illustrates overlapping stimulation patterns (electrode interaction) from the implanted electrodes as they stimulate neural regions.

Our group has been developing an image-guided cochlear implant programming (IGCIP) system [2], which uses image analysis techniques to assist audiologists with electrode interaction analysis and electrode configuration selection [18, 24, 36] during the CI programming procedure. Figure 2 shows the workflow of IGCIP. We use whole head computed tomography (CT) images of CI recipients as input for IGCIP. For recipients having both pre- and post-implantation CTs, we segment the intra-cochlear anatomy with the method described in [13]. In the post-implantation CT, electrodes are localized with methods described in [14-15], [27-28], where the method used

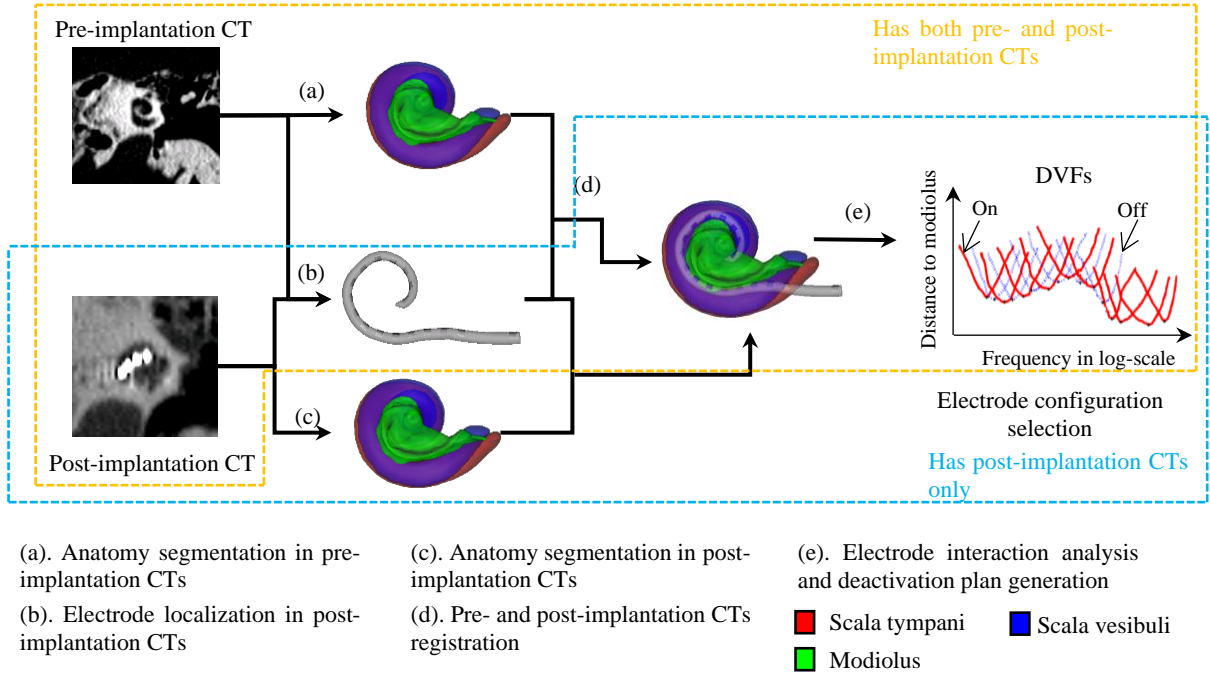

**Figure 2.** Workflow of Image-guided cochlear implant programming (IGCIP) techniques.

(a). Anatomy segmentation in pre-implantation CTs
(b). Electrode localization in post-implantation CTs
(c). Anatomy segmentation in post-implantation CTs
(d). Pre- and post-implantation CTs registration
(e). Electrode interaction analysis and deactivation plan generation

■ Scala tympani   ■ Scala vesibuli
■ Modiolus

depends on the model of the array. Then, we register the pre- and post-implantation CTs to analyze the possibility of electrode interactions. For recipients who do not have pre-implantation CTs, we developed two methods described in [16] and [17] that can segment the intra-cochlear anatomy directly from post-implantation CTs. After segmenting the intra-cochlear anatomy using one of these techniques, we localize the electrodes in the same post-implantation CTs by using automatic techniques developed by our group ([14-15], [27-28]) and then proceed to the electrode interaction analysis process. To analyze the electrode interactions, our group has developed a technique named distance-vs.-frequency curves (DVFs). The DVF is a 2D plot that captures the patient-specific spatial relationship between the electrodes and the auditory nerves [2], as shown in Figure 2. The DVFs show the distance from each electrode to neural stimulation sites along the length of the cochlea. Based on the DVFs, we have developed an automatic electrode configuration selection method [18] to select a subset of active electrodes that reduces electrode interaction. Recent clinical studies we have performed indicate that by using our IGCIP-generated electrode configuration, hearing outcomes can be significantly improved [19-21]. Because IGCIP uses the positions of the electrodes with respect to the anatomy, the electrode configuration it generates is affected by the accuracy of the anatomy segmentation and electrode localization techniques that are used. To better understand the sensitivity of IGCIP to these two steps in the process, we rigorously characterize them, and we study the effect that errors in these steps have on the overall process when considered individually or together. The results we have obtained allow us to draw conclusions on the accuracy of the algorithms we have developed and on the sensitivity of our programming suggestions to segmentation and localization errors.

The electrode localization method being evaluated in this study is a graph-based path-finding algorithm [14]. We refer to this method as $M_E$ (the subscript refers to electrode) in the remainder of this article. In post-implantation CTs, the CI electrodes appear as high intensity voxel groups, as shown in Figure 3. $M_E$ first extracts the volume of interest (VOI) that contains the cochlea by using a reference image. Next, it generates candidates of interest (COIs) that represent the potential locations of electrodes. The COIs are used as nodes in a graph. Then, it uses path-finding algorithms to find a path constructed by a subset of COIs representing the centroids of CI electrodes on the array. The intra-cochlear anatomy segmentation step in IGCIP focuses on the segmentation of three anatomical structures in the cochlea: the scala tympani (ST), the scala vestibuli (SV), and the active region (AR) of the modiolus (MOD). ST and SV are the two principal cavities of the cochlea. The MOD is the anatomical region housing the auditory nerves. AR is the interface between the MOD and the union of the ST and SV. The auditory nerves stimulated by the electrodes are located in the immediate proximity of AR within MOD. In conventional clinical pre-implantation CTs, the basilar membrane that separates ST and SV is not visible, as shown in Figure 3d, which makes the segmentation of the intra-cochlear anatomy difficult. When pre-implantation CTs are not available, the segmentation of the intra-cochlear anatomy becomes even more difficult. This is because in post-implantation CTs, the artifacts caused by metallic electrodes obscure the anatomical structures. Thus, for intra-cochlear anatomy segmentations in both pre- and post-implantation CTs, our group had proposed three automatic methods: (1) a statistical shape model-based method [13], (2) a library-based method [16], and (3) a method based on the Conditional Generative Adversarial Network (cGAN) [17]. We refer to them as $M_{A1}$, $M_{A2}$, and $M_{A3}$, respectively. $M_{A1}$ is used on pre-implantation CTs if available. In $M_{A1}$, we create an active shape model for ST, SV, and MOD by using manually delineated anatomical surfaces in 9 high resolution μCTs [13]. Then, the model is fit to the partial structures that are available in conventional CTs, and used to estimate the position of structures not visible in these CTs. When pre-implantation CTs are not available, we apply $M_{A2}$ or $M_{A3}$ directly to post-implantation CTs for intra-cochlear anatomy segmentation. $M_{A2}$ leverages a library of shapes of cochlear labyrinth and intra-cochlear anatomy. Given a target post-implantation CT, first, $M_{A2}$ segments the portions of the cochlear labyrinth that are not typically affected by image artifacts. Then, it selects a subset of labyrinth shapes from the library based on the similarity of the regions not affected by the artifacts. Using this subset of shapes, the method builds a weighted active shape model (wASM) of the cochlear labyrinth to localize the labyrinth in the target image. Then weights of the vertices that are close to (or distant to) the image artifacts are assigned 0 (or 1), respectively. Last, it uses another pre-defined active shape model of ST, SV, and MOD to segment the intra-cochlear anatomy based on the localized labyrinth. $M_{A3}$ uses a cGAN [17] that takes as input a post-implantation CT in which the intra-cochlear anatomy is corrupted by artifacts and synthesizes the corresponding pre-implantation artifact-free image. We then apply $M_{A1}$ to the synthesized image to generate the ST, SV and MOD surfaces.

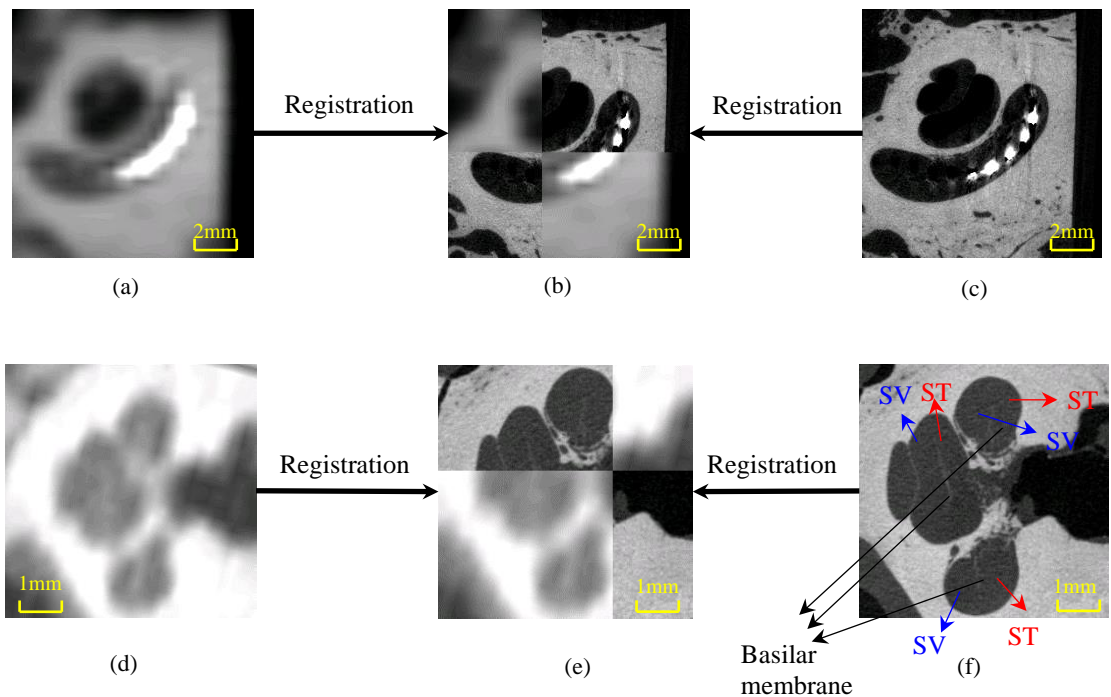

**Figure 3.** Panels a-c show three post-implantation CTs: a conventional CT (a), the registered μCT (c), and a checkerboard combination of the two (b). As can be seen, electrodes are more separable in the μCT because of the higher resolution and less partial volume artifacts. Panels d-f show three pre-implantation CTs: a conventional CT (d), the registered μCT (f), and a checkerboard combination of the two (e). As can be seen in panel (f) and (d), the basilar membrane is visible in μCTs but not visible in clinical CTs. This makes it possible for generating ground truth anatomy segmentation results for ST and SV, and then MOD.

Other researchers have investigated methods for CI electrode localization and intra-cochlear anatomy segmentations in clinical CTs. For CI electrode localization, Bennink et al. proposed a method [29] that utilizes the *a-priori* knowledge of the geometry of electrode arrays. Braithwaite et al. proposed a method [30] that uses spherical measures for electrode localization. Chi et al. proposed a deep learning-based method [34]. For intra-cochlear anatomy segmentation, Zhang et al. uses 3D U-Net [35] trained with limited ground truth data to segment the anatomy. Demarcy [31] used µCTs to model the variances in cochlea shape and proposed a joint shape and intensity model-based segmentation method. Gerber et al. [32] created statistical models for cochlea shapes and the variances of the insertion of CI electrode arrays by using a dataset consisting of CTs and µCTs. Kjer et al. [33] uses a library of temporal bone µCTs to construct a cochlear statistical deformation model. The model is further used for regularization of the non-rigid registration between a patient-specific CT and a μCT for patient-specific cochlear anatomy segmentation. However, it is difficult to compare the performance of all these methods above because they have all been evaluated on different private datasets owned by different groups, and there exists no standardized approach for the evaluation of the automatic techniques in the image-guided cochlear implant programming field. Thus, the goal of this study is to develop a dataset and a standardized approach that permit evaluating the sensitivity of IGCP with respect to electrode localization and anatomy segmentation algorithms. We demonstrate the use of our proposed procedure to evaluate several such algorithms that we have developed. However, in future studies the validation approach and the ground truth dataset being presented in this study could be used to similarly evaluate other methods such as ones developed by other groups.

As has been discussed above, to analyze the accuracy of IGCIP, we need to rigorously characterize the accuracy of the automatic image processing techniques. In previous studies, $M_E$, $M_{A2}$, and $M_{A3}$ have only been validated by using reference segmentation results on conventional CTs that have limited resolution (the typical voxel size in these volumes is 0.2×0.2×0.3mm3). In [14], to evaluate the accuracy of $M_E$, we used a set of manual localization results generated by an expert on post-implantation clinical CTs. When localizing small-sized objects such as CI electrodes (typical size is 0.3×0.3×0.1mm3), partial volume artifacts (see Figure 3a) in clinical CTs limit the accuracy of the localization, even when done with care by an expert. Other image quality issues, such as beam hardening artifacts, also complicate localizing CI electrodes. In previous studies performed to analyze intra-cochlear anatomy segmentation methods, $M_{A2}$ and $M_{A3}$ were only compared to $M_{A1}$ applied to corresponding pre-implantation CTs. The accuracy of the reference segmentations used in prior validation studies was thus limited by the resolution of clinical CT images that were used.

In this article, we create a high accuracy ground truth dataset using µCT images to rigorously evaluate the accuracy of the automatic techniques used in IGCIP and the sensitivity of the overall IGCIP process to segmentation and localization errors. In Section 2, we describe the creation of the ground truth dataset and the design of the validation approaches. In Section 3, we present and analyze the validation results. In Section 4, we summarize the contribution of this work and discuss potential improvements for the IGCIP process.

## 2. Methods

### 2.1 Image data

Our image data consists of CTs and µCTs of 35 temporal bone specimens implanted with 4 different types of CI electrode arrays by an experienced otologist. The detailed specifications of the 35 specimens are shown in Table 1. Among the 35 specimens, 20 (specimen 16 to 35 in Table 1) were implanted with an array type that our electrode localization method had been trained to localize, and the remaining 15 were implanted with three other array types (5 specimens each, specimen 1 to 15 in Table 1) for which our method was not trained. Every specimen underwent pre- and post-implantation CT imaging and post-implantation µCT imaging. Six specimens underwent pre-implantation µCT imaging (specimen 30 to 35). The typical voxel size for CT images and µCT images are $0.30 \times 0.30 \times 0.30 \text{mm}^3$ and $0.02 \times 0.02 \times 0.02 \text{mm}^3$, respectively.

### 2.2 Ground truth dataset creation

Figure 3 shows examples of pre- and post-implantation CTs and µCTs. As can be seen, the individual electrodes in a post-implantation µCT are more separable than in a conventional post-implantation CT because the µCT has 3 orders of magnitude better resolution and little partial volume artifact. The intra-cochlear anatomy is clearly more visible in a pre-implantation µCT than in a clinical CT. In particular, the basilar membrane is visible in a µCT when it is not in standard CTs. Delineating intra-cochlear structures is thus easier in µCTs.

**Table 1.** The specifications of the CT images of the 35 temporal bone specimens

| Specimen # | Conventional CT voxel size (mm3) | | µCT voxel size (mm3) | | Electrode migration | Data group # |
|---|---|---|---|---|---|---|
| | Pre-op CT | Post-op CT | Pre-op CT | Post-op CT | | |
| 1 | 0.26 × 0.26 × 0.30 | 0.26 × 0.26 × 0.30 | | 0.02 × 0.02 × 0.02 | | 1,3 |
| 2 | 0.28 × 0.28 × 0.30 | 0.24 × 0.24 × 0.30 | | 0.02 × 0.02 × 0.02 | | 1,3 |
| 3 | 0.30 × 0.30 × 0.30 | 0.34 × 0.34 × 0.30 | | 0.02 × 0.02 × 0.02 | Yes | 3 |
| 4 | 0.27 × 0.27 × 0.30 | 0.34 × 0.34 × 0.30 | | 0.02 × 0.02 × 0.02 | | 1,3 |
| 5 | 0.26 × 0.26 × 0.30 | 0.21 × 0.21 × 0.30 | | 0.02 × 0.02 × 0.02 | | 1,3 |
| 6 | 0.27 × 0.27 × 0.30 | 0.31 × 0.31 × 0.30 | | 0.02 × 0.02 × 0.02 | | 1,3 |
| 7 | 0.25 × 0.25 × 0.30 | 0.34 × 0.34 × 0.30 | | 0.02 × 0.02 × 0.02 | | 1,3 |
| 8 | 0.32 × 0.32 × 0.30 | 0.30 × 0.30 × 0.30 | | 0.02 × 0.02 × 0.02 | | 1,3 |
| 9 | 0.24 × 0.24 × 0.30 | 0.32 × 0.32 × 0.30 | | 0.02 × 0.02 × 0.02 | | 1,3 |
| 10 | 0.21 × 0.21 × 0.30 | 0.30 × 0.30 × 0.30 | | 0.02 × 0.02 × 0.02 | | 1,3 |
| 11 | 0.35 × 0.35 × 0.30 | 0.28 × 0.28 × 0.30 | | 0.02 × 0.02 × 0.02 | | 1,3 |
| 12 | 0.35 × 0.35 × 0.30 | 0.21 × 0.21 × 0.30 | | 0.02 × 0.02 × 0.02 | | 1,3 |
| 13 | 0.38 × 0.38 × 0.30 | 0.23 × 0.23 × 0.30 | | 0.02 × 0.02 × 0.02 | | 1,3 |
| 14 | 0.40 × 0.40 × 0.30 | 0.27 × 0.27 × 0.30 | | 0.02 × 0.02 × 0.02 | | 1,3 |
| 15 | 0.25 × 0.25 × 0.30 | 0.26 × 0.26 × 0.30 | | 0.02 × 0.02 × 0.02 | | 1,3 |
| 16 | 0.40 × 0.40 × 0.40 | 0.40 × 0.40 × 0.40 | | 0.03 × 0.03 × 0.03 | | 1,3 |
| 17 | 0.40 × 0.40 × 0.40 | 0.40 × 0.40 × 0.40 | | 0.03 × 0.03 × 0.03 | | 1,3 |
| 18 | 0.40 × 0.40 × 0.40 | 0.40 × 0.40 × 0.40 | | 0.03 × 0.03 × 0.03 | | 1,3 |
| 19 | 0.40 × 0.40 × 0.40 | 0.40 × 0.40 × 0.40 | | 0.03 × 0.03 × 0.03 | | 1,3 |
| 20 | 0.40 × 0.40 × 0.40 | 0.40 × 0.40 × 0.40 | | 0.03 × 0.03 × 0.03 | | 1,3 |
| 21 | 0.40 × 0.40 × 0.40 | 0.40 × 0.40 × 0.40 | | 0.03 × 0.03 × 0.03 | | 1,3 |
| 22 | 0.40 × 0.40 × 0.40 | 0.40 × 0.40 × 0.40 | | 0.03 × 0.03 × 0.03 | | 1,3 |
| 23 | 0.40 × 0.40 × 0.40 | 0.40 × 0.40 × 0.40 | | 0.03 × 0.03 × 0.03 | | 1,3 |
| 24 | 0.40 × 0.40 × 0.40 | 0.40 × 0.40 × 0.40 | | 0.03 × 0.03 × 0.03 | | 1,3 |
| 25 | 0.40 × 0.40 × 0.40 | 0.40 × 0.40 × 0.40 | | 0.03 × 0.03 × 0.03 | | 1,3 |
| 26 | 0.34 × 0.34 × 0.29 | 0.15 × 0.15 × 0.30 | | 0.02 × 0.02 × 0.02 | | 1,3 |
| 27 | 0.31 × 0.31 × 0.30 | 0.25 × 0.25 × 0.30 | | 0.02 × 0.02 × 0.02 | | 1,3 |
| 28 | 0.32 × 0.32 × 0.30 | 0.16 × 0.16 × 0.30 | | 0.02 × 0.02 × 0.02 | Yes | 3 |
| 29 | 0.30 × 0.30 × 0.30 | 0.20 × 0.20 × 0.30 | | 0.02 × 0.02 × 0.02 | Yes | 3 |
| 30 | 0.38 × 0.38 × 0.30 | 0.19 × 0.19 × 0.30 | 0.02 × 0.02 × 0.02 | 0.02 × 0.02 × 0.02 | Yes | 2,3 |
| 31 | 0.39 × 0.39 × 0.30 | 0.14 × 0.14 × 0.30 | 0.02 × 0.02 × 0.02 | 0.02 × 0.02 × 0.02 | | 1,2,3,4 |
| 32 | 0.33 × 0.33 × 0.30 | 0.20 × 0.20 × 0.30 | 0.02 × 0.02 × 0.02 | 0.02 × 0.02 × 0.02 | | 1,2,3,4 |
| 33 | 0.29 × 0.29 × 0.40 | 0.32 × 0.32 × 0.30 | 0.02 × 0.02 × 0.02 | 0.02 × 0.02 × 0.02 | | 1,2,3,4 |
| 34 | 0.32 × 0.32 × 0.30 | 0.23 × 0.23 × 0.30 | 0.02 × 0.02 × 0.02 | 0.02 × 0.02 × 0.02 | | 1,2,3,4 |
| 35 | 0.29 × 0.29 × 0.30 | 0.17 × 0.17 × 0.30 | 0.02 × 0.02 × 0.02 | 0.02 × 0.02 × 0.02 | Yes | 2,3 |

We use the dataset for four validation purposes: (1) Characterize the accuracy of the electrode localization method $M_E$. (2) Characterize the accuracy of the three existing intra-cochlear anatomy segmentation methods $M_{A1}$, $M_{A2}$, and $M_{A3}$. (3) Analyze the sensitivity of IGCIP with respect to the accuracy of the methods in (1) and (2). (4) Assess the quality of the IGCIP-generated electrode configurations generated with our complete automated process applied to clinical images. Using the images of the 35 specimens, we create 4 dataset groups and one "electrode configuration dataset". The 4 validation dataset groups are shown in Table 1. Details on each of these groups and on the electrode configuration dataset are provided in Section 2.3.

**2.3 Validation approaches**

**2.3.1 Error analysis for the electrode localization method**

We use Group 1 (see Table 1) to characterize the accuracy of $M_E$. It consists of 30 out of the 35 specimens with pre- and post-implantation CTs and post-implantation µCTs. Two experts manually determined the locations of electrodes in the post-implantation µCTs of these 30 specimens. We average the manual localizations from the two experts to generate the ground truth locations (GL) of the electrodes. The details for the GL generation process can be found in [23]. Then, we apply $M_E$ to the corresponding 30 conventional post-implantation CTs of the specimens in Group 1 to generate the automatic localization (AL) of the electrodes. Post-implantation conventional and µCTs were registered to facilitate comparison between automatic and gold-standard ground truth localizations using mutual information-based registration techniques. The registrations were visually inspected and confirmed to be accurate, as shown in Figure 3b. We do not include specimens 3, 28, 29, 30, and 35 in Group 1 because we observed that the CI electrode arrays had clearly moved between the conventional and the µCTs during visual inspection, which makes those 5 subjects not available for evaluating the accuracy of $M_E$. One example of specimen with electrode migration between post-implantation µCT and CT is shown in Figure 4a. We hypothesize that this motion occurred due to the fact that the specimen cochleae do not have fluid that could typically stabilize the array. Thus, when the specimens were transferred between different imaging sites, the electrode arrays were not internally fixed and may have moved. In addition to GL and AL, we also created an image-based localization (IL) as the average of multiple expert localizations in the CT images. To create IL, an expert manually generated electrode localization results for each case repeatedly until adding a new instance changes the position of each electrode in the average localization by no more than 0.05mm (approximately ¼ the width of a CT voxel). This indicated that the expert's localizations converged to the best localization manually achievable when using the conventional CTs. To compare two electrode localizations, we measured Euclidean distances between the centroids of the corresponding electrode points and compared AL and GL. However, the overall localization error is a combination of (1) algorithmic errors and what we refer to as (2) image-based errors. The algorithmic errors are caused by limitation of the automatic techniques. The image-based errors are caused by limitations in the quality of the conventional CTs. Thus, we compared IL and AL to estimate algorithmic errors. We also compared IL and GL to measure image-based errors. Results we have obtained for this study are presented in Section 3.1.

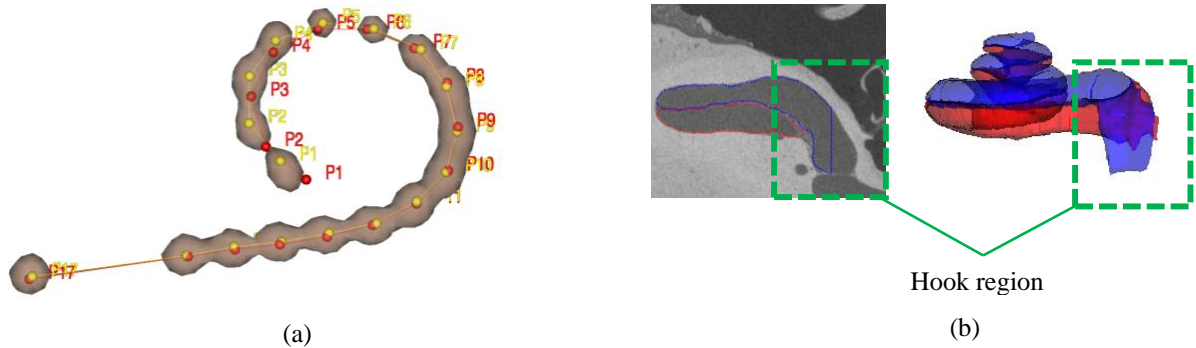

**Figure 4.** Panels (a) shows electrode migration in Specimen 3. The CT iso-surface of the highest intensity voxels is shown in orange. The automatically (yellow) and manually (red) localized electrodes from the CT and µCT are different from electrode P1 to P6. Panel (b) shows an axial slice of a µCT around the "hook region" of SV. The blue and red contours in the CT are the manual delineations of SV and ST generated by an expert. The corresponding 3D meshes are shown on the right side. As can be seen, the extent of the "hook region" of SV is chosen arbitrarily by the expert.

**2.3.2 Validation for intra-cochlear anatomy segmentation methods**

We use Group 2 (see Table 1) to evaluate the accuracy of the three intra-cochlear anatomy segmentation methods. Group 2 consists of 6 specimens for which post-implantation CTs, pre-implantation CTs, and pre-implantation µCTs are available. We apply $M_{A1}$ to the pre-implantation CTs, and $M_{A2}$ and $M_{A3}$ to the post-implantation CTs of the 6 specimens in Group 2, respectively. On the pre-implantation µCTs, an expert manually delineated the ST, SV, and MOD to serve as gold-standard ground truth for intra-cochlear anatomy. We registered pre-implantation and post-implantation CTs, and the pre-implantation µCTs together to facilitate the comparison of gold-standard segmentation results and automatic segmentation results. The automatic intra-cochlear anatomy segmentation methods generate surface meshes for ST, SV, and MOD that have pre-defined numbers of vertices. Those pre-defined numbers are different from the number of vertices in the manually generated surface meshes. To enable a point-to-point error estimation for manually and automatically generated meshes, we used an ICP-based [26] iterative non-rigid surface registration method developed in house to register the active shape model used to localize the ST, SV, and MOD to the manually delineated ST, SV, and MOD surfaces in the µCTs. This process results in a set of ground truth ST, SV, and MOD surfaces that have a one-to-one point correspondence with the surfaces generated by our automatic methods. For each intra-cochlear anatomy segmentation method, we then measured the Euclidean distance from each vertex on the automatically localized surfaces to the corresponding point on the gold-standard surfaces. The SV in the cochlea is a cavity with an open region on the side that is close to the round window membrane of the cochlea. In both CT and µCT, the border of the SV in the "hook region" (see Figure 4b) that is close to the round window membrane of the cochlea cannot be delineated consistently because the SV is an open cavity without an anatomical boundary at the hook region. Thus, the border must be estimated somewhat arbitrarily by the expert when generating the ground truth. Since the accuracy of the segmentation in this region is not important for intra-cochlear electrode localization or IGCIP, we exclude approximately 1.5cm$^3$ around the SV hook region when estimating the SV segmentation error. In the remainder of this article, we denote the gold-standard intra-cochlear anatomy surfaces as $S_0$, and the surfaces generated by using $M_{A1}$, $M_{A2}$, and $M_{A3}$ as $S_1$, $S_2$, and $S_3$. Accuracy results obtained with $M_{A1}$, $M_{A2}$, and $M_{A3}$ are presented in Section 3.2.

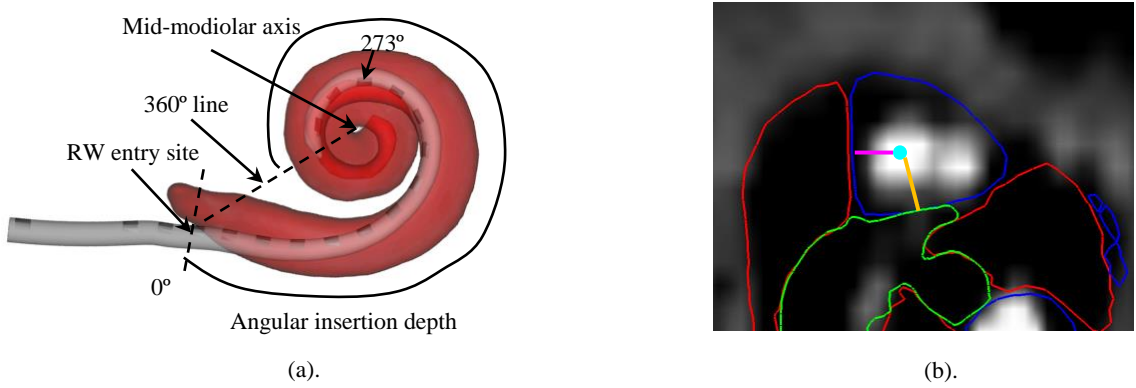

**Figure 5.** Panel (a) shows the measurement of the DOI value for the 3rd most apical electrode in the coordinate system proposed by Verbist et al. [25]. The ST is shown in red. The electrode array carrier is shown in light grey and the contacts are shown in dark grey. Panel (b) shows the measurements of DtoBM (magenta line) and DtoM (orange line) values for a given electrode (cyan point) in a CT slice in coronal view. The ST, SV and MOD are shown in red, blue, and green, respectively.

**2.3.3 Sensitivity of intra-cochlear electrode position estimation to processing errors**

We conduct three studies to analyze the effect of (a) localization errors, (b) segmentation errors, and (c) both segmentation and localization errors on the estimation of the position of contacts with respect to the inner ear anatomy. This is done with different groups of specimens as shown in Table 2 (study (a), (b), and (c)). As is shown in Figure 2, one electrode localization and one intra-cochlear anatomy segmentation define one estimation of the spatial relationship between the electrodes and auditory nerves. This relationship can be described by measuring locations of electrodes relative to intra-cochlear structures using an electrode coordinate system proposed by Verbist et al. [25]. As is discussed in Section 1, the intra-cochlear location of electrodes and their relationship to hearing outcomes has been a subject of intense study in recent years [5-10]. Thus, independently of IGCIP, it is of interest to quantify the accuracy of the processing methods for estimating intra-cochlear position to understand the limitations of these techniques for use in large scale studies assessing the effect of electrode position on outcomes. Thus, in this study, we quantify errors in estimating intra-cochlear electrode position when using $M_E$, $M_{A1}$, $M_{A2}$, and $M_{A3}$. Electrode position is measured in terms of angular depth-of-insertion (DOI), the distance to modiolar surface (DtoM), and the distance to the basilar membrane (DtoBM). As the cochlea has a spiral shape with 2.5 turns in humans, the depth of any position within it can be quantified in the terms of a DOI value from 0 to 900 degrees. The DtoM values are directly computed as the Euclidean distances between the centroids of electrodes and the vertices on the modiolar surface. The DtoBM value is computed as the signed Euclidean distance between the centroids of electrodes and the basilar membrane, which lies between ST and SV. Figure 5 show how these three values are computed. Among the three values, DOI and DtoM values are used to the construct the DVFs as they correspond to their horizontal and vertical axes. DtoM values are not directly related but still provide important information on the intra-cochlear locations of the implanted electrodes. The results we have obtained with this set of experiments are discussed in Section 3.3.

### 2.3.4 Sensitivity of IGCIP to processing errors

The spatial relationship between the electrodes and the intra-cochlear anatomy defines a set of DVFs. Based on the DVFs, an electrode deactivation plan, the "electrode configuration" is generated by using our automatic electrode configuration selection method [18]. In each study shown in Table 2, the sensitivity of IGCIP is defined as the difference between the electrode configurations generated when using the "automatic" and the "reference" intra-cochlear electrode position estimation. Table 2 defines the automatic and reference electrode position estimation techniques for each study and also provides the name we use for each resulting electrode configuration.

**Table 2.** Electrode configuration names in sensitivity analysis studies

| Study | Data group # | Intra-cochlear anatomy | Electrode locations | Configuration name |
|---|---|---|---|---|
| (a). Electrode localization sensitivity | 1 | $S_1$ | GL | $C_{G1}$ (Reference) |
| | | | AL | $C_{A1}$ |
| (b). Anatomy segmentation sensitivity | 2 | $S_0$ | GL | $C_{G0}$ (Reference) |
| | | $S_1$ | | $C_{G1}$ |
| | | $S_2$ | | $C_{G2}$ |
| | | $S_3$ | | $C_{G3}$ |
| | 3 | $S_1$ | GL | $C_{G1}$ (Reference) |
| | | $S_1'$ | | $C_{G1}'$ |
| | | $S_2'$ | | $C_{G2}'$ |
| | | $S_3'$ | | $C_{G3}'$ |
| (c). Overall sensitivity | 4 | $S_0$ | GL | $C_{G0}$ (Reference) |
| | | $S_1$ | AL | $C_{A1}$ |
| | | $S_2$ | AL | $C_{A2}$ |
| | | $S_3$ | AL | $C_{A3}$ |
| | 1 | $S_1$ | GL | $C_{G1}$ (Reference) |
| | | $S_1'$ | AL | $C_{A1}'$ |
| | | $S_2'$ | AL | $C_{A2}'$ |
| | | $S_3'$ | AL | $C_{A3}'$ |

In study (a), we evaluate the sensitivity of IGCIP with respect to the electrode localization method by using specimens in Group 1. The reference configurations in study (a) are defined as $C_{G1}$, which are generated by using $S_1$, together with GL. The automatic configurations are defined as $C_{A1}$, which are generated by using $S_1$ together with AL. In study (b), we evaluate the sensitivity of IGCIP with respect to the intra-cochlear anatomy segmentation methods by using specimens in Groups 2 and 3. In Group 2, which consists of the 6 subjects with pre-implantation μCTs, the reference configurations $C_{G0}$ are generated by $S_0$ together with the GL. The three sets of automatic configurations $C_{G1}$, $C_{G2}$, $C_{G3}$ are generated by using $S_1$, $S_2$, $S_3$ together with GL, respectively. Due to the limited number of pre-implantation μCTs acquired for subjects in our dataset, we use Group 3 to generate synthesized surfaces for $M_{A1}$, $M_{A2}$, and $M_{A3}$ so that we can analyze the sensitivity of IGCIP with respect to the errors introduced by the three intra-cochlear anatomy segmentation methods on a larger dataset. For the specimens in Group 3, we select $S_1$ of all the 35 specimens as our reference intra-cochlear anatomical surfaces. Then, for each subject, we deform $S_1$ to generate the synthesized surfaces $S_1'$, $S_2'$, $S_3'$ that simulate the segmentation errors of method $M_{A1}$, $M_{A2}$, and $M_{A3}$. To build synthesized surfaces $S_1'$ for $M_{A1}$, we first build a gamma distribution by using the mean and the standard deviation of the segmentation error of $M_{A1}$, which is estimated by using specimens in Group 2 and the error measurement approach described in sub-section 2.3.2. Then, for each specimen in Group 3, we draw a random number from the defined gamma distribution and set this number as the "desired mean segmentation error" between the synthesized surfaces and the reference surfaces of the selected subject. We randomly adjust the shape control parameters in the active shape model [22] so that we deform the reference surfaces to the synthesized surfaces with a mean point-to-point difference equal to the desired mean segmentation error. The same process is used to generate $S_2'$ and $S_3'$. We use an active shape model to perform this deformation, instead of directly adding errors to each vertex on the reference surface $S_1$, so that the changes in the deformed surfaces have realistic anatomical constraints. In Group 3, the reference configurations $C_{G1}$ are generated by using $S_1$ and GL. The three sets of automatic configurations $C_{G1}'$, $C_{G2}'$, $C_{G3}'$ are generated by using $S_1'$, $S_2'$, $S_3'$, together with GL, respectively. In study (c), we evaluate the sensitivity of IGCIP with respect to both the electrode and anatomy segmentation methods by using specimens in Group 4 and 1. Group 4 consists of the 4 specimens that have pre-implantation μCTs and do not have electrode migration. The reference configurations $C_{G0}$ in Group 4 in study (c) are generated by using the anatomy $S_G$, together with the GL. The three sets of automatic configurations $C_{A1}$, $C_{A2}$, and $C_{A3}$ are generated by using $S_1$, $S_2$, $S_3$, together with AL, respectively. Due to the same issue with the limited pre-implantation µCTs in study (b), for study (c), we use Group 1, which consists of the 30 specimens that do not have electrode migration to expand the size of our dataset for overall sensitivity analysis. The reference configurations $C_{G1}$ in Group 1 are generated by using $S_1$ and GL. The three sets of automatic configurations $C'_{A1}$, $C'_{A2}$, $C'_{A3}$ are generated by using $S'_1$, $S'_2$, $S'_3$, together with AL, respectively.

The most direct way to show the difference between two electrode configurations is to use a binary code (use "1" to indicate an electrode being "activated" and "0" to indicate an electrode being "deactivated") to represent the two configurations and then compute the hamming distance between them. This directly shows the differences between two given configurations. However, sometimes a configuration of "on-off-on-off-on" produces a stimulation pattern equivalent to the pattern produced by a "off-on-off-on-off", even though they result in large hamming distance. Thus, we use two other metrics to compare the automatic and reference configurations to evaluate the sensitivity of IGCIP. The first metric we use is the difference between "cost values" of the two configurations. In our automatic electrode deactivation strategy [18], we have developed a cost function which assigns a cost value to a specific electrode configuration. In our design, a lower cost value indicates a configuration that is less likely to cause electrode interaction and more likely to stimulate a broad frequency range. Thus, the difference between the cost values of two configurations is an indicator of the difference between the automatic and the reference electrode configurations. The second metric is the difference between the subjective quality of the automatic and reference electrode configurations. The quality of the electrode configurations is evaluated by an expert (JHN) through an electrode configuration quality assessment study. In previous studies [23-24], we found no statistically significant difference when comparing the ratings of the electrode configuration quality across multiple experts. Thus, in this study, we selected the most experienced expert for the quality assessment task. The details of the quality assessment study are discussed in the next subsection.

**2.3.5 Electrode configuration quality assessment**

We evaluate the quality of all the automatic electrode configurations in all three studies listed in Table 2 in one expert evaluation experiment. In study (a), (b), and (c), there are in total 255 automatic configurations. Each automatic configuration is compared to the reference configuration for the corresponding specimen as well as a control configuration. The quality of each configuration is determined by an expert using the reference DVFs for the specimen. Note that the reference DVFs, generated based on the reference electrode and anatomy localizations, will in general be different from the DVFs used to generate the automatic configuration. The quality of the automatic configuration when applied to the reference DVFs compared to the reference configuration represents the sensitivity of IGCIP to the automatic processing methods used to create the automatic configuration. The control configuration for each case was manually selected by one expert (YZ) to be a configuration that is not acceptable but close to acceptable. It is included to minimize the risk for the rater to be biased toward evaluating all configurations as acceptable and as a means to detect this bias if it exists.

An electrode configuration is judged as "acceptable" when the expert believes it can be used for CI programming and is likely to lead to acceptable hearing outcomes. The 255 automatic configurations, together with their reference and control configurations form 255 electrode configuration sets. The order of the 255 sets and the orders of the three configurations in each set are randomly shuffled. Each set is then presented to the human rater who is blinded to the identity of each configuration. For each set the three configurations are ranked. Each configuration is also labelled as acceptable or not.

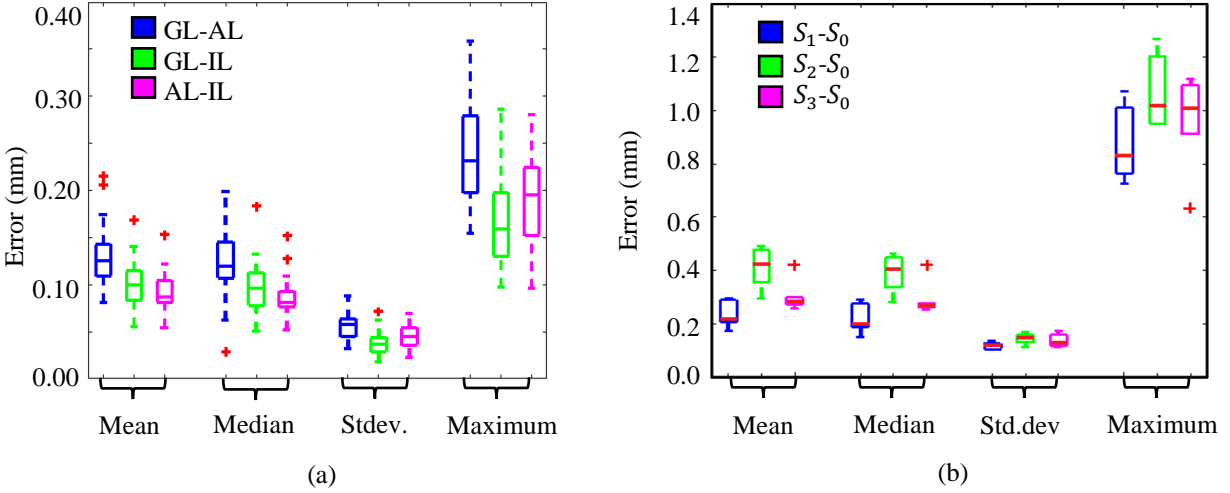

**Figure 6.** Panel (a) shows the boxplots for localization errors between AL-GL, IL-GL, and AL-IL. Panels (b) shows the segmentation errors between $S_1$-$S_0$, $S_2$-$S_0$, and $S_3$-$S_0$.

## 3. Results

### 3.1. Accuracy of the electrode localization technique

Validation of the electrode localization technique was presented in [23], and the results are summarized here. Figure 6a shows boxplots of the mean, median, maximum, and the standard deviation of localization errors between AL and GL across the 30 specimens in Group 1. In each boxplot, the median value is shown as a red line, the 25th and 75th percentiles are indicated by the blue box, whiskers show the range of data points that fall within 1.5x the interquartile range from the 25th or 75th percentiles but are not considered outliers, and red crosses indicate outlier data points. Comparing AL and GL, we found a mean electrode localization errors of 0.13mm and a maximum localization error of 0.36mm. Comparing IL and GL, we found a mean electrode localization error of 0.10mm and the maximum localization error of 0.29mm. Comparing AL and IL, we found mean and maximum localization errors equal to 0.09mm and 0.28mm, respectively. This shows that our automatic method generated localization results close to the optimal localization results that can be generated by an expert from clinical post-implantation CTs. All localization errors were smaller than the length of one voxel diagonal of the conventional post-implantation CTs in our dataset. The mean localization errors of AL-GL and IL-GL are comparable. We performed a paired t-test between the mean localization errors between AL-GL and IL-GL and found a p-value of $p < 10^{-6}$. Even though this result indicates that the AL and IL localization still have significant difference, the mean errors between AL-GL (0.13mm) and IL-GL (0.10mm) are very close and AL can generate results that are nearly as desirable as the most accurate manual localization results achievable by two experts from the clinical CTs. We also performed a paired t-test between the mean localization errors between AL-GL and AL-IL and found a p-value of $p < 10^{-8}$. This indicates that the ground truth manually generated by the expert on µCTs (GL) are significantly different from the ground truth manually generated by the expert on clinical CTs (IL), even though they have a small mean difference of 0.10mm, as discussed above. Thus, if post-implantation µCTs are available, we prefer to use GL to evaluate the electrode localization method. If µCTs are not available, IL can be used as an acceptable substitute of GL for estimating the localization errors of an electrode localization method.

**3.2. Accuracy of intra-cochlear anatomy segmentation methods**

Figure 6b show the boxplots of the mean, the maximum, the median, and the standard deviation of the differences between automatic segmentation methods and the ground truth across the 6 specimens in Group 2. Comparing $S_0$ and $S_1$, the mean and standard deviation of the segmentation errors was 0.23±0.12mm. Comparing $S_0$ and $S_2$, the mean and the standard deviation of the segmentation errors was 0.41±0.15mm. Comparing $S_0$ and $S_3$, the mean and the standard deviation of the segmentation errors was 0.30±0.14mm. Finally, among the three existing automatic methods in IGCIP and our gold-standard ground truth, we found the most accurate method was $M_{A1}$. This is because $M_{A1}$ uses pre-implantation CTs in which the metallic artifacts caused by electrodes do not exist. $M_{A3}$ results in better mean segmentation errors than $M_{A2}$ on post-implantation CTs. Overall, all three methods had <0.5mm mean segmentation errors. Figure 7 shows the segmentations of ST, SV, and AR from one case generated by all the methods. The surfaces are color-coded by using the segmentation errors computed by using $S_0$.

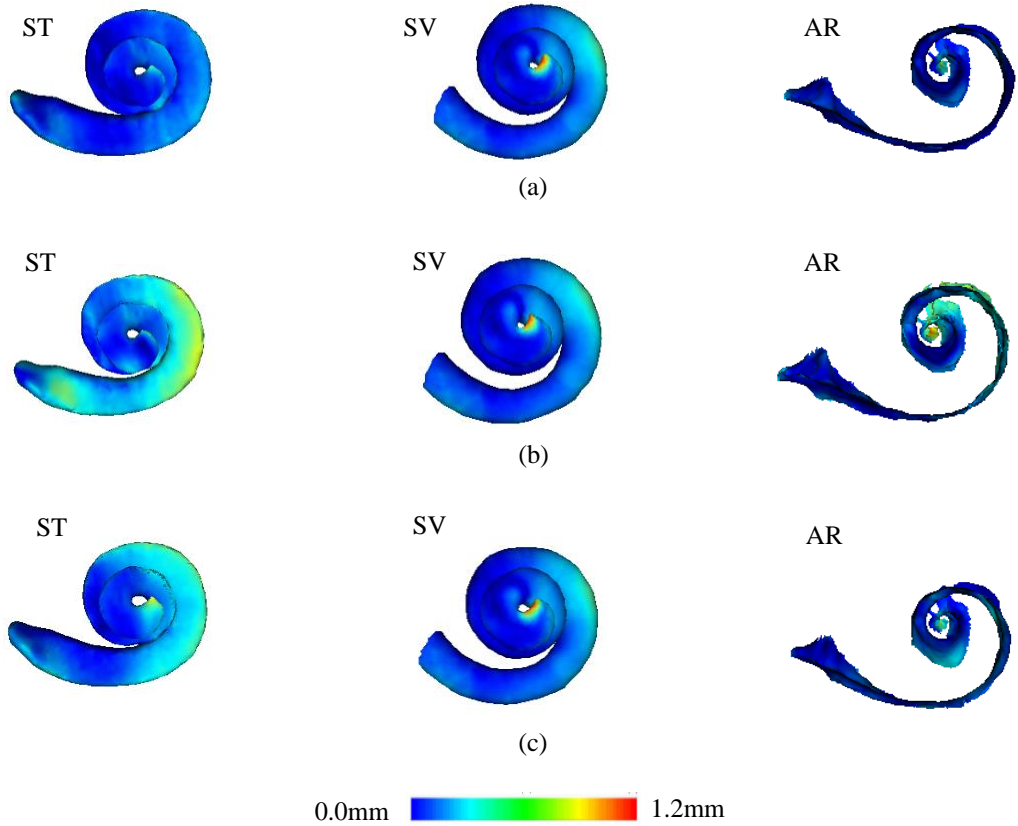

**Figure 7.** Panels (a), (b), (c) show qualitative segmentation results ($S_1$, $S_2$, and $S_3$) generated by IGCIP automatic methods $M_{A1}$, $M_{A2}$, and $M_{A3}$ for a representative subject in Group 2. The three surfaces of intra-cochlear anatomical structures are color-coded by the segmentation errors computed by using $S_0$.

**3.3 Sensitivity of intra-cochlear electrode position estimation to processing errors**

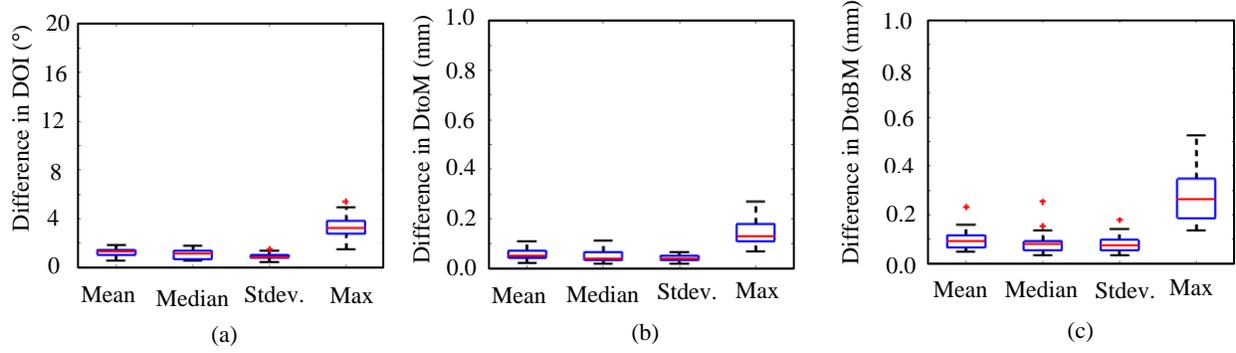

**Figure 8.** Panels (a), (b), and (c) show the boxplots for the differences in the DOIs, the DtoM, and the DtoBM of the automatic ($C_{A1}$) and the reference ($C_{G1}$) configurations generated by IGCIP for sensitivity analysis with respect to the electrode localization method (study (a) in Table 2).

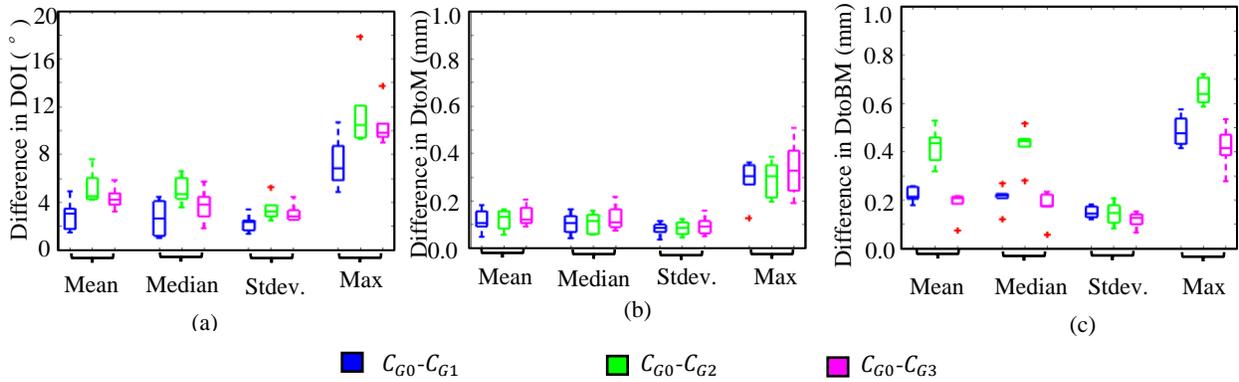

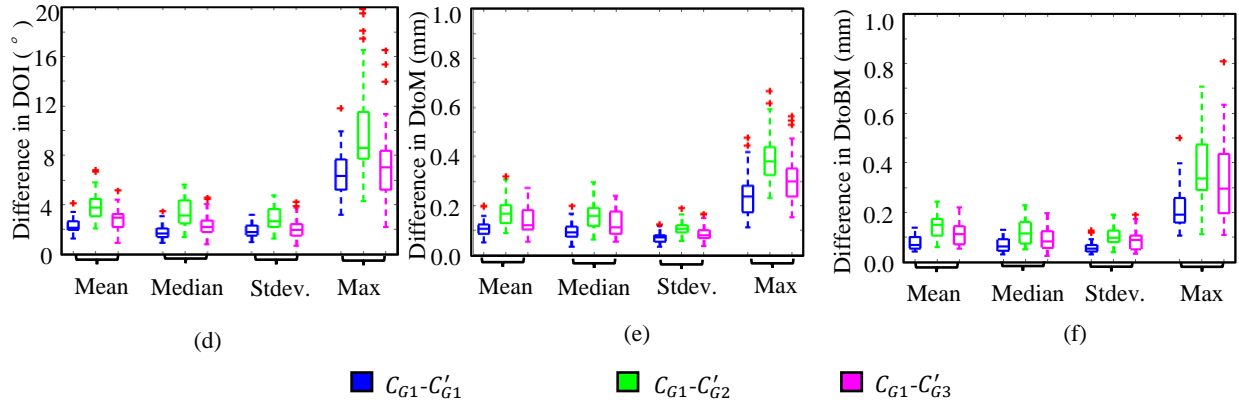

**Figure 9.** Panels (a-c) show the boxplots for the differences in the DOIs, the DtoM, and the DtoBM of the electrodes generated by using automatic ($C_{G1}, C_{G2}, C_{G3}$) and the reference ($C_{G0}$) processing methods on the 6 specimens in Group 2. Panels (d-f) show the boxplots for the differences in the DOIs, the DtoM, and the DtoBM of the electrodes generated by using automatic ($C'_{G1}, C'_{G2}, C'_{G3}$) and the reference ($C_{G1}$) processing methods on the 35 specimens in Group 3 with the synthesized anatomy surfaces. These results relate to the IGCIP sensitivity analysis study with respect to the intra-anatomy segmentation method (study (b) in Table 2).

**Table 3.** p-values of t-test results on the difference in DOIs, DtoM and DtoBM of $C'_{G1}$-$C_{G1}$, $C'_{G2}$-$C_{G1}$, $C'_{G3}$-$C_{G1}$

|  | $C'_{G1}$-$C_{G1}$ | | | $C'_{G2}$-$C_{G1}$ | | | $C'_{G3}$-$C_{G1}$ | | |
|---|---|---|---|---|---|---|---|---|---|
| Measurements | DOI | DtoM | DtoBM | DOI | DtoM | DtoBM | DOI | DtoM | DtoBM |
| $C'_{G1}$-$C_{G1}$ | / | / | / | 7.00e-7 | 1.48e-7 | 4.40e-8 | 6.40e-3 | 1.35e-2 | 3.32e-4 |
| $C'_{G2}$-$C_{G1}$ | | | | / | / | / | 1.34e-4 | 3.30e-3 | 2.68e-2 |
| $C'_{G3}$-$C_{G1}$ | | | | | | | / | / | / |

**Table 4.** p-values of t-test results on the difference in DOIs, DtoM and DtoBM of $C'_{A1}$-$C_{G1}$, $C'_{A2}$-$C_{G1}$, $C'_{A3}$-$C_{G1}$

|  | $C'_{A1}$-$C_{G1}$ | | | $C'_{A2}$-$C_{G1}$ | | | $C'_{A3}$-$C_{G1}$ | | |
|---|---|---|---|---|---|---|---|---|---|
| Measurements | DOI | DtoM | DtoBM | DOI | DtoM | DtoBM | DOI | DtoM | DtoBM |
| $C'_{A1}$-$C_{G1}$ | / | / | / | 1.14e-5 | 7.80e-8 | 4.26e-8 | 1.72e-3 | 5.00e-3 | 1.60e-3 |
| $C'_{A2}$-$C_{G1}$ | | | | / | / | / | 1.60e-3 | 3.20e-3 | 2.70e-3 |
| $C'_{A3}$-$C_{G1}$ | | | | | | | / | / | / |

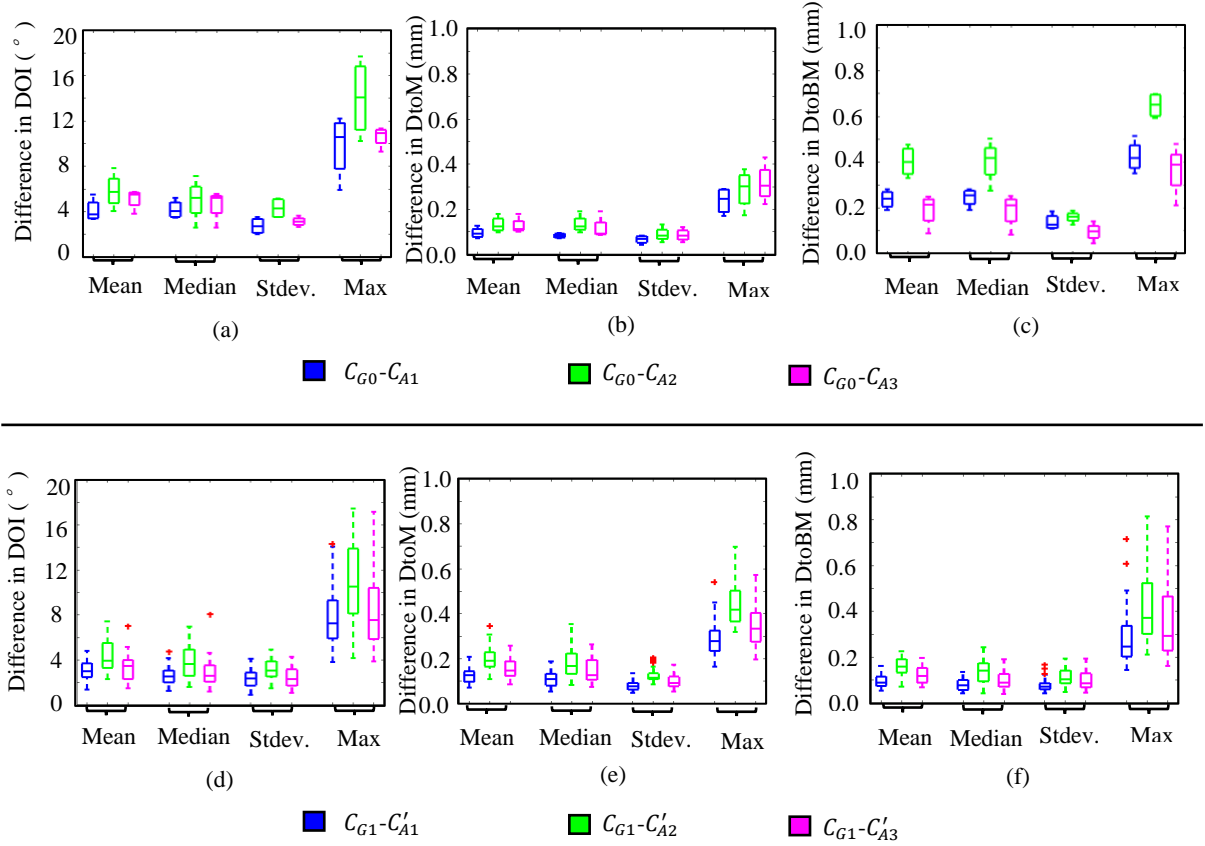

**Figure 10.** Panels (a-c) show the boxplots for the differences in the DOIs, the DtoM, and the DtoBM of the electrodes generated by using automatic ($C_{A1}$, $C_{A2}$, $C_{A3}$) and the reference ($C_{G0}$) processing methods on the 4 specimens in Group 4. Panels (d-f) show the boxplots for the differences in the DOIs, the DtoM, and the DtoBM of the electrodes generated by using automatic ($C'_{A1}$, $C'_{A2}$, $C'_{A3}$) and the reference ($C_{G1}$) processing methods on the 30 specimens in Group 1 with the synthesized anatomy surfaces. These are the results of the IGCIP sensitivity analysis study with respect to the overall process (study (c) in Table 2).

Figures 8-10 show boxplots for the difference between the intra-cochlear locations of the electrodes identified by using the automatic and the reference processing methods defined in study (a), (b), and (c) in Table 2. Comparing the results presented in Figure 8 and Figure 9, we find that the intra-cochlear locations of the electrodes are less sensitive to the electrode localization method than to the intra-cochlear anatomy segmentation methods. Among the three intra-cochlear anatomy segmentation methods, $M_{A1}$ is the most reliable method for generating accurate intra-cochlear locations, then $M_{A3}$, followed by $M_{A2}$. Comparing the results presented in Figure 8-10, we find that the overall errors of both the electrode localization and intra-cochlear anatomy segmentation

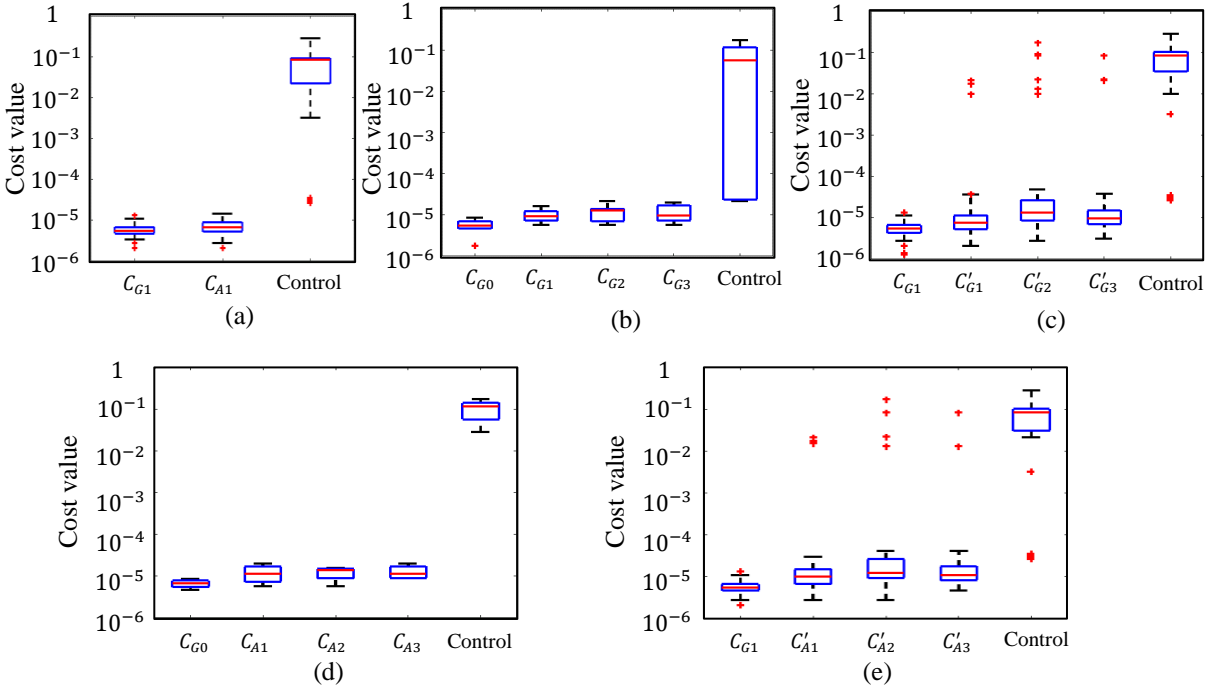

**Figure 11.** Panels (a-e) show the boxplots for the cost values (in log-scale) of automatic, reference, and control configurations for subjects in the data being used in the three studies in Table 2 for IGCIP sensitivity analysis.

techniques are not substantially larger than the errors due to the intra-cochlear anatomy segmentation alone.

By using the synthesized anatomical surfaces, we perform a paired t-test on the difference of the average three measurements (DOI, DtoM, and DtoBM) of the intra-cochlear electrode locations of the automatic and reference configurations generated by using the electrode locations and anatomical surfaces in study (b) and (c). With Bonferroni correction, we found that $M_{A1}$, $M_{A2}$, and $M_{A3}$ lead to significantly different intra-cochlear electrode locations. More specifically, as can be seen from Figure 9d-f and Figure 10d-f, the mean error of the DOI, DtoM and DtoBM values generated by using $M_{A1}$ are significantly lower than the ones generated by using $M_{A2}$ and $M_{A3}$.

### 3.4. Sensitivity of IGCIP to processing errors

Table 5. p-values of t-test results among the cost values of $C'_{G1}$, $C'_{G2}$, $C'_{G3}$, and $C_{G1}$ (reference).

|  | $C_{G1}$ (Reference) | $C'_{G1}$ | $C'_{G2}$ | $C'_{G3}$ |
|---|---|---|---|---|
| $C_{G1}$ (Reference) | / | 9.50e-3 | 1.68e-4 | 3.40e-3 |
| $C'_{G1}$ |  | / | 3.78e-2 | 3.30e-1 |
| $C'_{G2}$ |  |  | / | 7.62e-2 |
| $C'_{G3}$ |  |  |  | / |

Table 6. p-values of t-test results on the cost values of $C'_{A1}$, $C'_{A2}$, $C'_{A3}$, and $C_{G1}$ (reference)

|  | $C_{G1}$ (Reference) | $C'_{A1}$ | $C'_{A2}$ | $C'_{A3}$ |
|---|---|---|---|---|
| $C_{G1}$ (Reference) | / | 7.40e-3 | 9.73e-4 | 4.60e-3 |
| $C'_{A1}$ |  | / | 3.78e-1 | 9.65e-1 |
| $C'_{A2}$ |  |  | / | 1.04e-1 |
| $C'_{A3}$ |  |  |  | / |

In Figure 11, we show the boxplots for the cost values of the automatic, reference, and control configurations defined in sub-section 2.3.5. The name of the configurations are indexed in Table 2. From Figure 11, we can see that besides the outliers, the average cost values for all the automatic configurations are close to the average cost values for the reference configurations. The average cost values for the control configurations are substantially larger than the ones for the reference and the automatic configurations. These results show that the automatic image processing techniques in our IGCIP can generate configurations whose cost values that are similar to the cost values of configurations generated by using the reference anatomy and electrode locations. From Figure 11a, we see that $M_E$ generates electrode locations that lead to cost values similar to the ground truth electrode locations. The p-value of the paired t-test on the cost values of $C_{G1}$ and $C_{A1}$ is $7.67 \times 10^{-2}$. This indicates we do not find a significant difference between the cost values of $C_{G1}$ and $C_{A1}$. From Figure 11b-e, we see that $M_{A1}$ generates the intra-cochlear anatomy that leads to a lower average cost than $M_{A2}$ and $M_{A3}$. This is because $M_{A1}$ is applied on pre-implantation CTs, where the intra-cochlear anatomy is not obscured by the metallic artifacts. For the two methods designed for post-implantation CTs, $M_{A3}$ generates the intra-cochlear anatomy that leads to lower average cost than $M_{A2}$. This indicates that $M_{A3}$ is more reliable than $M_{A2}$. This is also shown in the differences in the DOI and the DtoBM values in Figure 9 and Figure 10. Table 5 and 6 show the p-values of the results of the cost values of the automatic and reference configurations presented in Figure 11c and 11e, respectively. From these p-values in Table 5, we see that the cost values of the configurations generated by $C_{G1}$ are significantly different than the ones generated by $C'_{G1}$, $C'_{G2}$, and $C'_{G3}$. However, there is no significant difference among the cost values of the configurations generated by $C'_{G1}$, $C'_{G2}$, and $C'_{G3}$. The p-values in Table 6 also support a similar conclusion regarding $C'_{A1}$, $C'_{A2}$, and $C'_{A3}$. Statistical analysis on the cost values of the configurations demonstrates that the differences in the intra-cochlear locations of electrodes generated by $M_{A1}$, $M_{A2}$, $M_{A3}$ lead to significantly different electrode configurations by IGCIP in terms of the cost values when compared with the electrode configurations generated by using reference intra-cochlear locations of electrodes. The p-values of $C_{G1} - C'_{G1}$ and $C_{G1} - C_{A1}$ show that the errors in our intra-cochlear anatomy localization method significantly affect the cost function value of the electrode configurations selected for IGCIP, whereas the electrode localization method does not lead to configurations with significantly different cost values compared to the ground truth.

Figure 12 shows results of the qualitative evaluation for the 255 electrode configuration sets in our electrode configuration dataset discussed in sub-section 2.3.5. In Figure 12, panel (a) shows the evaluation results of the configurations generated for the sensitivity analysis of IGCIP with respect to the electrode localization method. These configurations belong to study (a) in Table 2. Panel (b) and (c) show the evaluation results of the configurations generated for the sensitivity analysis of IGCIP with respect to the three intra-cochlear anatomy segmentation methods. These configurations belong to study (b). Panel (d) and (e) show the evaluation results of the configurations generated for the overall sensitivity analysis of IGCIP with respect both the electrode and anatomy segmentation methods for study (c). As can be seen in Figure 12a, among the 30 automatic electrode configurations in $C_{A1}$ generated by using AL, none of them in is rated as not acceptable, and 21 out of 30 automatic configurations in $C_{A1}$ are rated as at least equally as good as the reference configurations $C_{G1}$. This shows that the electrode localization method is robust enough to generate localization results that lead to acceptable electrode deactivation configurations.

In Figure 12b, among the automatic configurations $C_{G1}$, $C_{G2}$, and $C_{G3}$ generated by using GL and $S_1$, $S_2$, and $S_3$, none is rated as not acceptable. Meanwhile, 4, 3, and 2 of $C_{G1}$, $C_{G2}$, and $C_{G3}$, respectively, are rated as at least equally as good as the reference configurations $C_{G0}$. In Figure 12c, among the automatic configurations $C'_{G1}$, $C'_{G2}$, and $C'_{G3}$ generated by using GL and $S'_1$, $S'_2$, $S'_3$; 2, 8, and 3 are rated as not acceptable; and 26, 14, and 15 are rated as at least equally good as the reference configurations $C_{G1}$. The results shown in Figure 12a-c show that the quality of the IGCIP-generated electrode configurations generated are less sensitive to the errors in the electrode localization method than to the intra-cochlear anatomy segmentation methods. In Figure 12d, among the automatic configurations $C_{A1}$, $C_{A2}$, and $C_{A3}$ generated by using AL and $S_1$, $S_2$, $S_3$, none of them is rated as unacceptable. Three of $C_{A1}$ are rated as equally as good as the reference configurations $C_{G0}$. In Figure 12e, among the automatic configurations $C'_{A1}$, $C'_{A2}$, and $C'_{A3}$ generated by using AL and $S'_1$, $S'_2$, $S'_3$; 4, 10, and 5 are rated as not acceptable; and 17, 11, and 14 are rated as at least as good as the reference configurations $C_{G1}$. Altogether, these results suggest that $M_{A1}$ is the most reliable anatomy localization method to generate acceptable electrode configurations. Further, $M_{A3}$ should be used as the secondary choice for anatomy segmentation when pre-implantation CTs are not available and $M_{A1}$ cannot be directly used. For statistical analysis, we performed McNemar mid-$p$ test on the acceptance rate of different groups of configurations presented in Figure 12c and 12e. The p-values of these analyses are shown in Table 7 and 8. As can be seen, none of the group of the configurations has significantly different acceptance rate than the others. Combining the statistical test results shown in Table 3-4 and Table 4-5 with the McNemar mid-$p$ test results on the acceptance rates shown in Table 7 and 8, after applying the Bonferroni correction, we find that the three anatomy segmentation methods are not significantly different from each other in terms of the quality of configurations generated by using them. This is also showing that even though the three anatomy segmentation methods $M_{A1}$, $M_{A2}$, $M_{A3}$ lead to significantly different intra-cochlear locations for the implanted electrodes, those differences are trivial and the three methods still lead to configurations with no significant difference in quality.

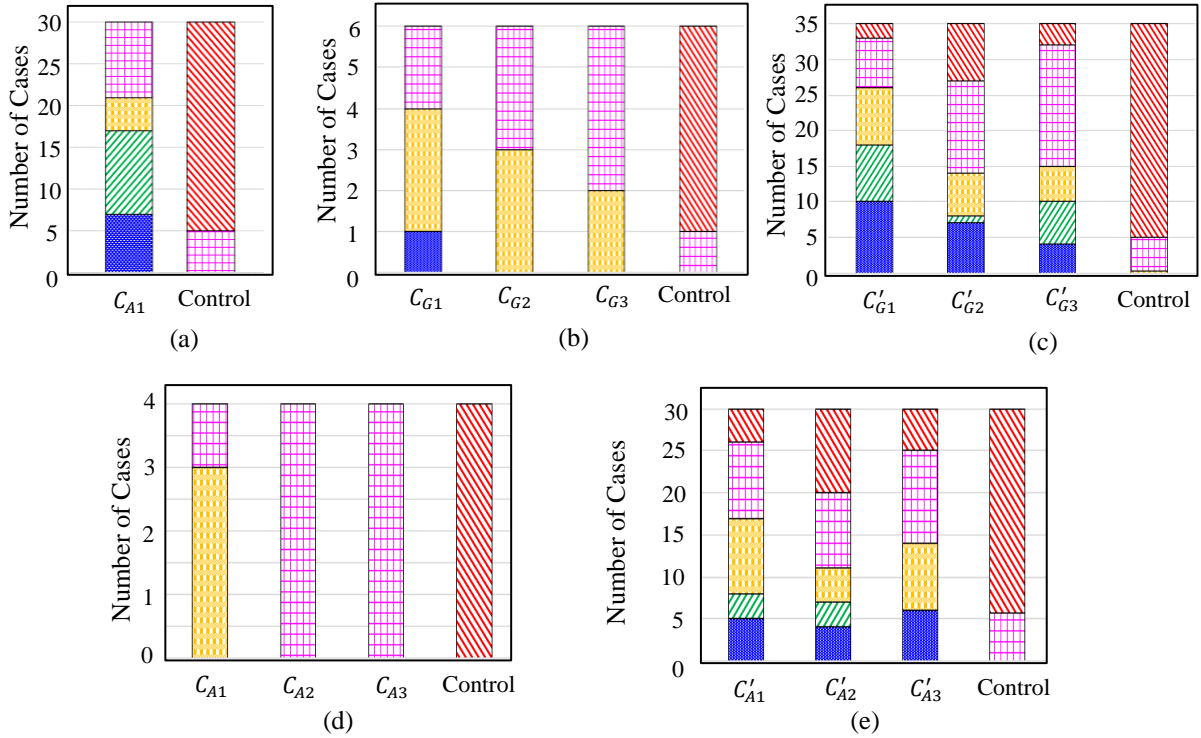

**Figure 12.** Evaluation results of the configurations generated for the sensitivity analysis of IGCIP with respect to (a) the electrode localization method, (b-c) the three intra-cochlear anatomy segmentation methods, and (d-e) the overall automatic image processing techniques in IGCIP.

In the results shown in Figure 12e, the expert evaluated 26 out of 30 (86.7%) automatic configurations generated by $M_E + M_{A1}$ as acceptable, and 25 out of 30 (83.3%) automatic configurations generated by $M_E + M_{A3}$ as acceptable. These results, together with the results presented in Section 3.3, indicate that when applied to clinical CT images, our image processing methods lead to electrode configuration recommendations that are reliable in the great majority of the cases. Our results also indicate that to further improve the reliability of IGCIP, we should aim to increase the accuracy of the intra-cochlear anatomy segmentation methods.

In Figure 12a-e, we show that among all the control configurations in all the experiments, 83.3%, 83.3%, 85.7%, 100%, and 81.1% are rated as unacceptable by the expert. This suggests that the evaluation results generated by the expert shown above are not biased towards a tendency to rate every configuration as acceptable.

## 4. Conclusion

In this article, we create a highly accurate ground truth dataset and a validation approach for the evaluation of automatic techniques for image-guided cochlear implant programming. Using the dataset and the validation approach, we perform a validation study on an image-guided cochlear implant programming (IGCIP) system we have developed. The two major image processing steps in our IGCIP system are CI electrode localization and intra-cochlear anatomy segmentation. The validation study results we have obtained show that among 30 cases in our dataset, our localization method can generate results that are highly accurate with mean and maximum electrode localization errors of 0.13mm and 0.36mm, respectively. Our three intra-cochlear anatomy localization methods can generate results that have mean errors of 0.23mm, 0.41mm, and 0.30mm. In a sensitivity analysis for IGCIP, we found that IGCIP is less sensitive to the electrode localization method than to the intra-cochlear anatomy segmentation method. Among the three intra-cochlear anatomy segmentation methods, we found that IGCIP is least sensitive to method $M_{A1}$, then $M_{A3}$, followed by $M_{A2}$. In an overall electrode configuration quality evaluation study, we found that IGCIP can generate configurations that are 86.7% acceptable when the pre-implantation CTs are available, and 83.3% acceptable when the pre-implantation CTs are not available. The validation approach and the ground truth dataset can also be applied for the evaluation of other image processing techniques proposed by other groups in image-guided cochlear implant programming. One limitation of this study is that while it includes several models of CI electrode arrays, they were produced by only one manufacturer. In the future, we plan to expand the validation dataset by acquiring pre- and post-implantation CTs and µCTs of temporal bone specimens implanted with electrode arrays from different CI manufacturers. We will also study hearing outcomes of CI recipients using IGCIP-generated configurations and the manually selected configurations to show the effectiveness of IGCIP-generated configurations.

**Table 7.** p-values of McNemar test results on the acceptance rate of $C'_{G1}$, $C'_{G2}$, $C'_{G3}$

|  | $C'_{G1}$ | $C'_{G2}$ | $C'_{G3}$ |
|---|---|---|---|
| $C'_{G1}$ | / | 1.13e-1 | 1.00 |
| $C'_{G2}$ |  | / | 1.82e-1 |
| $C'_{G3}$ |  |  | / |

**Table 8.** p-values of McNemar test results on the acceptance rate of $C'_{A1}$, $C'_{A2}$, $C'_{A3}$

|  | $C'_{A1}$ | $C'_{A2}$ | $C'_{A3}$ |
|---|---|---|---|
| $C'_{A1}$ | / | 1.48e-1 | 1.00 |
| $C'_{A2}$ |  | / | 2.28e-1 |
| $C'_{A3}$ |  |  | / |

## Acknowledgements

This work was supported in part by grants R01DC014037 and R01DC014462 from the National Institute on Deafness and Other Communication Disorders. We would also like to thank Advanced Bionics, LLC (Valencia, CA) for providing materials used in this study. The content is solely the responsibility of the authors and does not necessarily represent the official views of these institutes.

## Disclosure of conflicts of interests

Dr. Robert F. Labadie is a consultant with Advanced Bionics and Ototronix.

# References

[1]. National Institute on Deafness and Other Communication Disorders. NIDCD Fact Sheet: Cochlear Implants. 2011  Available from: http://www.nidcd.nih.gov/staticresources/health/hearing/FactSheetCochlearImplant.pdf.

[2]. Noble, J.H., et al., Image-guidance enables new methods for customizing cochlear implant stimulation strategies. Neural Systems and Rehabilitation Engineering, IEEE Transactions on, 2013. 21(5): pp. 820-829.

[3]. Gifford, R.H., A. Hedley-Williams, and A.J. Spahr, Clinical assessment of spectral modulation detection for adult cochlear implant recipients: a non-language based measure of performance outcomes. Int J Audiol, 2014. 53(3): pp. 159-64.

[4]. Gifford, R.H., J.K. Shallop, and A.M. Peterson, Speech recognition materials and ceiling effects: considerations for cochlear implant programs. Audiol Neurootol, 2008. 13(3): pp. 193-205.

[5]. Aschendorff, A., et al., Quality control after cochlear implant surgery by means of rotational tomography. Otology & Neurotology, 2005. 26(1): pp. 34-37.

[6]. Rubinstein, J.T., How cochlear implants encode speech. Current opinion in otolaryngology & head and neck surgery, 2004. 12(5): pp. 444-448.

[7]. Skinner, M.W., et al., In vivo estimates of the position of advanced bionics electrode arrays in the human cochlea. Annals of Otology, Rhinology & Laryngology, 2007. 116(4): pp. 2-24.

[8]. Verbist, B.M., et al., Multisection CT as a valuable tool in the postoperative assessment of cochlear implant patients. American Journal of Neuroradiology, 2005. 26(2): pp. 424-429.

[9]. Wanna, G.B., et al., Impact of electrode design and surgical approach on scalar location and cochlear implant outcomes. The Laryngoscope, 2014. 124(S6): pp. S1-S7.

[10]. Wanna, G.B., et al., Assessment of electrode placement and audiologic outcomes in bilateral cochlear implantation. Otology & neurotology: official publication of the American Otological Society, American Neurotology Society [and] European Academy of Otology and Neurotology, 2011. 32(3): pp. 428.

[11]. Boex, C., de Balthasar, C., Kos, M. I., and Pelizzone, M. Electrical field interactions in different cochlear implant systems, The Journal of the Acoustical Society of America, vol. 114, pp. 2049–2057, 2003.

[12]. Fu, Q.J., and Nogaki, G., Noise susceptibility of cochlear implant users: The role of spectral resolution and smearing, Journal of the Association for Research in Otolaryngology, vol. 6, no. 1, pp. 19–27, 2005.

[13]. Noble, J.H., et al., Automatic Segmentation of Intra-Cochlear Anatomy in Conventional CT. IEEE Trans Biomed Eng, 2011. 58(9): p. 2625-32.

[14]. Zhao, Y., Chakravorti, S., Labadie, R.F., Dawant, B.M., and Noble, J.H., Automatic graph-based method for localization of cochlear implant electrode arrays in clinical CT with sub-voxel accuracy. Medical Image Analysis, 52, 2019. pp 1-12. doi: 10.1016/j.media.2018.11.005

[15]. Zhao, Y., Dawant, B.M, Labadie, R.F., and Noble, J.H., Automatic localization of closely spaced cochlear implant electrode arrays in clinical CTs. Medical Physics. doi: 10.1002/mp.13185

[16]. Reda, F.A., Noble, J.H., Labadie, R.F., and Dawant, B.M., An artifact-robust, shape library-based algorithm for automatic segmentation of inner ear anatomy in post-cochlear-implantation CT. Proceedings of SPIE--the International Society for Optical Engineering, 2014. 9034: p. 90342V.

[17]. Wang, J., Zhao, Y., Noble, J.H., Dawant, B.M., Conditional Generative Adversarial Networks for metal artifact reduction in CT Images of the head. MICCAI 2018 Sep; 11070:3-11. doi: 10.1007/978-3-030-00928-1_1.

[18]. Zhao, Y., Dawant, B.M., and Noble, J.H., Automatic selection of the active electrode set for image-guided cochlear implant programming, Journal of Medical Imaging, 3(3), 035001 (2016), doi: 10.1117/1.JMI.3.3.035001.

[19]. Noble, J.H., Gifford, R.H., Hedley-Williams, A.J., Dawant, B.M., Labadie, R.F., Clinical Evaluation of an Image-Guided Cochlear Implant Programming Strategy. Audiol. & Neurotol. 2014; 19: 400-411.

[20]. McRackan, T.R., Noble, J.H., Wilkinson E.P., Mills D., Dietrich M.S., Dawant B.M., Gifford R.H., and Labadie R.F., Implementation of Image-Guided Cochlear Implant Programming at a Distant Site. Otolaryngol. Head Neck Surg., 2017; 156 (5): 933-937.

[21]. Noble, J.H., Hedley-Williams A.J., Sunderhaus L., Dawant B.M., Labadie R.F., Camarata S.M., Gifford R.H., Initial results with Image-guided Cochlear Implant Programming in Children. Otol. Neurotol., 2016; 37 (2): e63-9.

[22]. Cootes, T.F., Taylor, C.J., Cooper, D.H., and Graham, J., Active Shape Models-Their Training and Application. Computer Vision and Image Understanding, 1995. 61(1): p. 38-59.

[23]. Zhao, Y., Labadie, F.R., Dawant, J.H., and Dawant, M.B., Validation of automatic cochlear implant electrode localization techniques using μCTs, Journal of Medical Imaging, 5(3), 035001 (2018). doi: 10.1117/1.JMI.5.3.035001.

[24]. Zhang, D., Zhao, Y., Noble, J.H., and Dawant, B.M., Selecting electrode configurations for image-guided cochlear implant programming using template matching. Journal of Medical Imaging, 5(2), 021202 (2018), doi: 10.1117/12.2256000.

[25]. Verbist, B.M., Skinner, M.W., Cohen, L.T., Leake, P.A., James, C., Boëx, C., Holden, T.A., Finley C.C., Roland, P.S., Roland, J.T. Jr., Haller, M., Patrick, J.F., Jolly, C.N., Faltys, M.A., Briaire, J.J., Frijns, J.H., Consensus panel on a cochlear coordinate system applicable in histologic, physiologic, and radiologic studies of the human cochlea, Otol. and Neurotol., 31(5), 722-730, 2010.

[26]. Besl, P.J., McKay, N.D., A method for registration of 3D shapes, IEEE Trans. On Pattern Analysis and Machine Intelligence, 14(2), 1992.

[27]. Zhao, Y., Dawant, B.M., Labadie, R.F., and Noble, J.H., Automatic localization of cochlear implant electrodes in CT. in Medical Image Computing and Computer-Assisted Intervention–MICCAI 2014. 2014, Springer. p. 331-338.

[28]. Zhao, Y., Dawant, B.M., and Noble, J.H., Automatic localization of cochlear implant electrodes in CTs with a limited intensity range. Proceedings of SPIE--the International Society for Optical Engineering, 2017. 10133-101330T.

[29]. Bennink, E., Peters, J.P.M., Wendrich, A.W., Vonken, E., van Zantan, G.A., and Viergever, M.A., 2017. Automatic localization of cochlear implant electrode contacts in CT. Ear and Hearing. 38(6): e376-e384. doi: 10.1097/AUD.0000000000000438.

[30]. Braithwaite, B., Kjer, H.M., Fagertun, J., Ballaster, M.A.G., Dhanasingh, A., Mistrik, P., Gerber, N., Paulsen, R.R., 2016. Cochlear implant elecrode localization in post-operative CT using a spherical measure. Biomedical Imaging (ISBI), 2016 IEEE 13th International Symposium on, Prague, Czech Republic.

[31]. Demarcy, T., Segmentation and study of anatomical variability of the cochlea from medical images. Universite Cote d'Azur, 2017.

[32]. Gerber, N., Reyes, M., Barazzetti, L., Kjer, H.M., Vera, S., Stauber, M., Mistrik, P., Ceresa, M., Mangado, N., Wimmer, W., Stark, T., Paulsen, R.R., Weber, S., Caversaccio, M., and Gonzalez Ballester, M.A., Data Descriptor: A multiscale imaging and modeling dataset of the human inner ear. Science Data: 4, 170132 (2017).

[33]. Kjer, H.M., Fagertun, J., Wimmer, W., Gerber, N., Vera, S., Barazzetti, L., Mangado, N., Ceresa, M., Piella, G., Stark, T., Stauber, M., Reyes, M., Weber, S., Caversaccio, M., Ballester, M.A.G., Paulsen, R.R., Int J CARS (2018) 13: 389. https://doi.org/10.1007/s11548-017-1701-7

[34]. Chi, Y., Wang, J., Zhao, Y., Dawant, B.M., Noble, J.H., A deep-learning-based method for the localization of cochlear implant electrodes in CT images. 2019 IEEE 16th International Symposium on Biomedical Imaging (ISBI 2019), Venice, Italy, 2019, pp. 1141-1145. doi: 10.1109/ISBI.2019.8759536

[35]. Zhang, D., Banalagay, B., Wang, J., Zhao, Y., Noble, J.H., Dawant, B.M., Two-level training of a 3D U-Net for accurate segmentation of the intra-cochlear anatomy in head CTs with limited ground truth training data, Proc. SPIE 10949, Medical Imaging 2019: Image Processing, 1094907. https://doi.org/10.1117/12.2512529

[36]. Zhao, Y., Dawant, B.M., Noble, J.H., Automatic electrode configuration selection for image-guided cochlear implant programming, Proc. SPIE 9415, Medical Imaging 2015: Image-Guided Procedures, Robotic Interventions, and Modeling, 94150K. https://doi.org/10.1117/12.2081473